\documentclass[conference]{IEEEtran}
\IEEEoverridecommandlockouts

\usepackage{cite}
\usepackage{amsmath,amssymb,amsfonts}
\usepackage{graphicx}
\usepackage{textcomp}
\usepackage{xcolor}
\usepackage[T1]{fontenc}
\usepackage[utf8]{inputenc}
\usepackage{xurl}
\usepackage[hidelinks]{hyperref}
\usepackage{booktabs}
\usepackage{stfloats}
\def\BibTeX{{\rm B\kern-.05em{\sc i\kern-.025em b}\kern-.08em
    T\kern-.1667em\lower.7ex\hbox{E}\kern-.125emX}}

\newcommand{\code}[1]{\texttt{#1}}
\newcommand{\blind}[2]{\ifblind #1\else #2\fi}

\newif\ifblind
\blindfalse 

\begin{document}

\title{\vspace*{0.25in}RescueVLM: Model-Agnostic Vision-Language Framework for Rescue Drone Imagery\\
\ifblind\else
\thanks{The code, including all system prompts, is publicly available at \protect\url{https://github.com/RoblabWh/RescueVLM}.\protect\\[\baselineskip]
This work is funded by the Federal Ministry of Research, Technology and Space (BMFTR) under grant 13N16478, project ``E-DRZ: Establishment of the German Rescue Robotics Center'' (DRZ) \protect\url{https://www.rettungsrobotik.de}.}
\fi
}

\ifblind
\author{\IEEEauthorblockN{Anonymous Author(s)}
\IEEEauthorblockA{Details omitted for double-blind review}}
\else
\author{
\IEEEauthorblockN{\small Niklas Voigt, Tim Sibum, Julien Meine, Niklas Digakis, Hartmut Surmann}
\IEEEauthorblockA{\textit{Robotics Laboratory} \\
\textit{Westphalian University of Applied Sciences (WHS)} \\
Gelsenkirchen, Germany \\
\small \{niklas.voigt, julien.meine, niklas.digakis, hartmut.surmann\}@w-hs.de, \\
\small tim.sibum@studmail.w-hs.de}
}
\fi

\maketitle

\begin{abstract}
A single unmanned aerial vehicle (UAV) mission over an incident site produces hundreds of images, far
more than a rescue commander can review under time pressure, and human
screening of destruction imagery carries a documented psychological cost. The
two obvious remedies both fail in search and rescue (SAR) practice: cloud
vision-language model (VLM) services are excluded because the imagery must not
leave the incident network under GDPR and German civil-protection (BOS) rules,
and fine-tuned specialist detectors age quickly while re-training and
re-certifying them exceeds most SAR organizations' resources. We present
RescueVLM, an open-source, training-free and model-agnostic triage tool built
on SketchVLM: a general-purpose open-weight VLM runs fully on-premise and is
swapped for a newer model through a simple dropdown selection, so triage
quality grows with the underlying models because all task knowledge lives in
prompts and an output contract rather than in trained weights. One
natural-language query over a mission folder returns a ranked shortlist with
per-image justifications, editable SVG overlay annotations, and a GPS map of
the hits. In a qualitative evaluation on German fire-brigade field data from
three real deployments and a live fire exercise, a single query condenses an entire survey mission into a
handful of annotated frames that a commander can review at a glance.
\end{abstract}


\section{Introduction}
\label{sec:intro}

\begin{figure}[!t]
\centering
\includegraphics[width=\columnwidth]{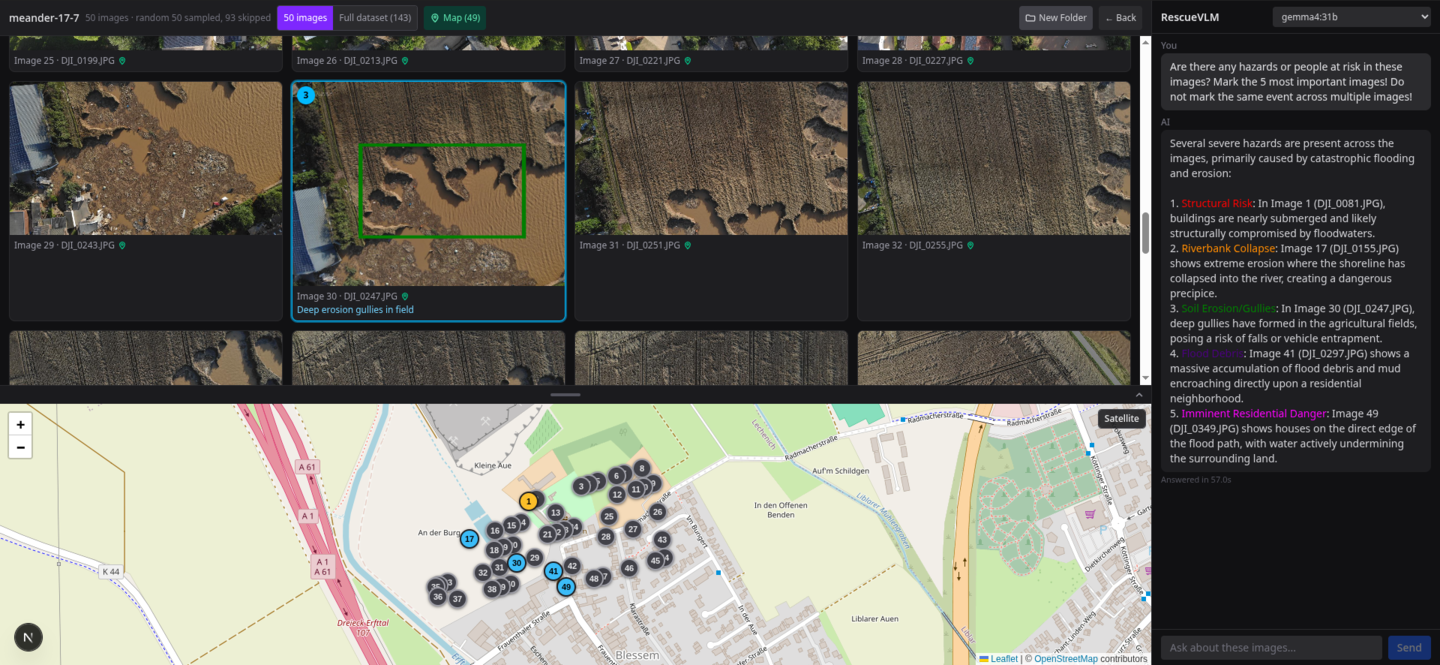}
\caption{RescueVLM triage run on the flood use case, Erftstadt/Blessem 2021
(fast mode): image grid with ranked hits, EXIF-GPS map, and situation summary.
The hits flag severe flood- and erosion-related hazards: nearly submerged,
likely structurally compromised buildings, a collapsed riverbank
forming a dangerous precipice, deep gullies in agricultural fields
posing fall and vehicle-entrapment risk, flood debris and mud
encroaching on a residential area, and houses at the edge of the
flood path where water is undermining the surrounding ground.}
\label{fig:flood}
\end{figure}

Unmanned aerial vehicles (UAVs) have become a primary sensor for situational awareness
in disaster response~\blind{\cite{fernandes2018harvey}}{\cite{surmann2021indoor,fernandes2018harvey}}. Mapping and survey flights produce a plain folder of hundreds of
georeferenced JPEGs per mission; at Hurricane Harvey, 112 flights generated roughly
150~GB of imagery and video across the deployment~\cite{fernandes2018harvey}.
\blind{During the July 2021 Erftstadt/Blessem flood deployment, the bottleneck
was never capturing imagery but getting current, condensed information
to the commanders who had to act on
it~\cite{surmann2022erftstadt,surmann2024lessons}.}{In our own deployment
during the July 2021 Erftstadt/Blessem flood, the bottleneck was never
capturing imagery but getting current, condensed information to the
commanders who had to act on
it~\cite{surmann2022erftstadt,surmann2024lessons}.}
This is the problem RescueVLM addresses: turning a folder of survey images into
a short, explained, map-anchored shortlist a commander can act on in minutes
(Fig.~\ref{fig:flood}).

Three constraints make this problem specific to SAR, and together they rule
out the obvious off-the-shelf solutions. The first is privacy. SAR imagery is
personal data of the most sensitive kind: victims, injured or deceased
persons, private interiors exposed by structural collapse, license plates.
Under the GDPR and German BOS practice, uploading such
imagery to a foreign cloud API during a live operation is de facto excluded;
local, on-premise inference is a requirement, not an optimization. Yet most
recent VLM-for-UAV systems assume exactly the cloud frontier model that such
deployments cannot use (Sec.~\ref{sec:related}).

The second constraint is that fine-tuned specialists do not survive in
practice. A task-specific detector or grounding
model~\cite{wang2026locateanything} is exemplary the day it ships and
superseded the day a better fine-tune appears, and for a SAR organization each
replacement means new training data, training, validation, and, for
operational BOS use, acceptance testing. Incident imagery is moreover rare,
privacy-restricted, and heterogeneous (the datasets in this paper span
vegetation, structural, industrial, and flood incidents)\blind{}{, and our own earlier fine-tuned
SAR detector~\cite{surmann2023uavs} shows precisely this treadmill}. The
durable artifact in this domain is the framework around the model, not
the model.

The third constraint is the psychological cost of human review. Today, teams
of humans screen post-disaster imagery frame by frame, and this exposure is
documented to cause harm, from operator stress when viewing victim imagery at
the World Trade Center~\cite{murphy2014disaster} to PTSD symptoms among military drone operators~\cite{chappelle2014drone}; a NOAA cartographer among the
first to screen post-Katrina aerial imagery recalls sitting at her desk in
tears~\cite{noaa2019aerial}. A pre-triage that surfaces only the handful of
images that matter reduces the volume of content that operators would need to review~\cite{hughes2025seeing}.

RescueVLM answers all three constraints at once. We rebuild
SketchVLM~\cite{collins2026sketchvlm}, a training-free, model-agnostic
framework in which a VLM explains its answers by emitting editable SVG
overlays on the input image, for on-site SAR use: whole mission folders are
queried in bulk against a local Ollama~\cite{ollama2026software} backend, with
\code{gemma4:31b}~\cite{googledeepmind2026gemma4} on a single workstation GPU,
no cloud dependency, and no fine-tuned weights. Because the model is reached through
a standard API, a better open-weight VLM is a drop-in replacement and the tool
inherits the improvement the same day.

We make three contributions.
\begin{enumerate}
  \item RescueVLM, an open-source adaptation of SketchVLM for fully local,
    training-free triage of SAR UAV imagery, into which any vision-capable
    open-weight model drops as a replaceable engine.
  \item A bulk folder mode that ranks $N$ images per natural-language query
    with per-image justifications, normalized SVG annotations, an explicit
    cross-image deduplication instruction for 70--80\,\% meander overlap, and
    ranked hits geo-anchored on a map via EXIF-GPS.
  \item A qualitative use-case evaluation on German fire-brigade field data
    (vegetation fire, structural fire, hazardous-substance reconnaissance, and
    flood) on responder-grade single-GPU hardware. 
\end{enumerate}

\section{Related Work}
\label{sec:related}
A fast-growing line of work applies vision-language models to aerial disaster
imagery. UAV-VLRR~\cite{yaqoot2025uavvlrr} and
UAV-VLA~\cite{sautenkov2025uavvla} use GPT-class cloud models for scene
interpretation and text-to-mission planning;
DisasTeller~\cite{chen2026disasteller} is a multi-LVLM agent framework for
post-disaster damage assessment and reporting, and RAPID~\cite{yang2026rapid}
is, like us, a training-free damage-assessment pipeline. Seeing the
Storm~\cite{hughes2025seeing} uses multimodal LLMs to filter overwhelming
disaster media so managers view only relevant items, a direct precedent for
our exposure-reduction thesis. RescueVLM differs from all of these on three
axes at once: it runs fully on-premise (cloud systems cannot be deployed on
GDPR-restricted incident imagery), it treats the model as a swappable engine
behind an output contract, and it targets bulk set-level triage-ranking with
cross-image deduplication rather than per-image captioning or classification.

A second line fine-tunes specialist grounding and detection models. Open-set
detectors from Grounding DINO~\cite{liu2024groundingdino} to the current
LocateAnything-3B~\cite{wang2026locateanything} produce sharp boxes on the
categories they were trained for, but their value lives in the weights, which
is exactly what the next fine-tune replaces; none of them rank a set, justify
a decision in text, or deduplicate events across a meander flight. We treat
such models as replaceable engines a framework can call, not as the system.

On the human-factors side, and beyond the evidence of screening harm cited in
Sec.~\ref{sec:intro}, RescueVLM inverts the screening workflow so that far fewer people see
far fewer disturbing frames while the human remains the decision maker, a
design informed by the appropriate-reliance literature~\cite{lee2004trust}.

Finally, that a local open-weight VLM can be ``good enough'' is increasingly
supported~\cite{deitke2025molmo}, and on-premise deployment for GDPR-sensitive
data has direct precedent in clinical text~\cite{wiest2024privacy}. Two
reliability caveats shape our design: off-the-shelf VLMs hallucinate
objects~\cite{li2023pope}, and forcing structured (JSON) output can degrade
reasoning~\cite{tam2024speakfreely}, which is why triage ordering, not
diagnosis, is the tool's contract.

\section{System Description}
\label{sec:system}

RescueVLM inherits the core paradigm of SketchVLM~\cite{collins2026sketchvlm}:
a training-free VLM tool that explains its answer by emitting a
non-destructive, editable overlay on the input image. We rebuild it for
privacy-preserving, on-site triage of survey imagery. The tool is a web application
that analyzes folders of mission images. In this mode, the responder points the tool
at a folder of mission JPEGs and asks one natural-language question across the whole image set.


The interface supports a quick mode that randomly samples up to
50 images from a given folder and a full-dataset mode that processes larger collections
sequentially in batches of up to 50 images. The 50-image limit per batch is imposed by GPU VRAM constraints. Before
submission, each image is resized in the browser to a maximum
longest side of 1024\,px and re-encoded as JPEG. Images are preceded
by labels of the form \code{"Image N: filename (GPS: lat, lon)"},
with GPS coordinates included when available. Image indices remain
consistent across batches and the interface.

Each batch is analysed independently, producing a short summary and a list of relevant images with consistent image numbers, brief explanations, and visual annotations. Once all batches have been processed, the same model receives the original question and the collected findings, including filenames and available GPS coordinates, in a final text-only request. It is instructed to identify repeated observations of the same event, compare their severity, and produce a combined answer and an overall image ranking. This step uses the batch findings without viewing the images again. The selected images retain their annotations from the batch analyses. If the final synthesis fails, the individual batch results are displayed together; failed batches are explicitly reported. Follow-up questions use the collected text findings and conversation history without resending the images.

A single thin server route proxies the browser to an OpenAI-compatible
\code{/v1/chat/completions} endpoint and streams the reply back as server-sent
events. The provider is selected by the model tag: tags prefixed
\code{ollama:} are routed to a local Ollama~\cite{ollama2026software} server, so imagery never leaves the incident network;
routes to remote providers exist in the code but stay disabled during
operations. The interface queries the local Ollama tag list and offers every
installed vision model in a dropdown, with no
change to code, data, or weights. In our deployment configuration the engine is
\code{gemma4:31b}~\cite{googledeepmind2026gemma4} on a single high-end GPU,
validated on a 32\,GB RTX 5090.

All task knowledge reaches the model as text, on two levels. The system prompt
defines the output contract: the reply must be strict JSON of the form
\code{\{answer, images:[\{index, note, annotations[]\}]\}}, where \code{answer}
is the situation summary, \code{images} lists relevant images
most-relevant-first with the first entry treated as the top pick, \code{note}
is a short (about ten words) justification, and every returned image must
carry at least one annotation marking the region that justified its selection.
Annotations use a normalized 0--1000 coordinate system relative to each image
and a fixed primitive set (\code{number}, \code{text}, \code{circle},
\code{rect}, \code{path} with SVG \code{M}/\code{L}/\code{Q}/\code{C}/\code{Z}
commands, \code{arrow}, and \code{polygon}), and key phrases in \code{answer}
are wrapped in colored spans matching the annotation they refer to, linking
the textual and the visual explanation. The system prompt also instructs the
model to examine every image before answering instead of stopping at the first
plausible match. The task semantics live in the user query: the canonical
triage prompt of Sec.~\ref{sec:eval}, offered in the interface as a one-click
suggestion, asks for the five most important images and instructs the model
not to mark the same event across multiple frames, the deduplication that
70--80\,\% meander overlap makes necessary. Adapting the tool to a new task,
or fixing an unwanted behavior, is an edit to this text rather than a training
run.

Because open-weight models occasionally violate the contract, the reply passes
through a fault-tolerant parser that strips markdown fences, recovers the
outermost JSON object by brace matching, and accepts common malformations such
as image entries given as bare numbers, so a single formatting slip does not
discard a whole triage result; if all recovery fails, the raw text is still
shown as a plain answer. Annotations are rendered as an SVG layer over the 1024\,px image that the model processes. Because the coordinates are resolution independent, the
overlay scales with the image: when the responder enters the editing view to
pose a follow-up question, the image is shown at 2048\,px and the marks are
scaled accordingly. There, the model's marks are pre-loaded as ordinary
editable shapes the responder can move, extend, or erase. GPS positions are
extracted fully client-side by a dependency-free EXIF reader that parses the
GPS records from the first 256\,KB of each JPEG, and geotagged hits are placed
on a map with rank-colored, numbered pins carrying the triage note (Fig.~\ref{fig:flood}). Map tiles
are the one online dependency, so fully offline operation additionally
requires a local tile cache; the imagery itself never leaves the machine.

\section{Evaluation}
\label{sec:eval}
We evaluate the way the tool
is used: for each use case we state the commander's question, show what the
system returned on the corresponding deployment data, and discuss what it found
and what it missed. All runs use the identical
framework and output contract of Sec.~\ref{sec:system} with \code{gemma4:31b}
served by a local Ollama instance on a single Nvidia RTX 5090. 50 images are processed in under a minute. Unless
stated otherwise, the query is the canonical triage prompt: it asks for
hazards and people at risk, requests the five most important images, and
forbids marking the same event across multiple images.

\subsection{Field data and use cases}
\label{sec:data}
We evaluate on raw, unlabeled image folders from deployments of the German Rescue Robotics Center (DRZ) TaskForce: the
folder of georeferenced JPEGs that lands
on the commander's laptop is the operational input distribution, including
oblique frames, 70--80\,\% meander overlap, and many images that contain nothing
of interest. All imagery was captured by DJI
enterprise UAVs, during or after real incidents except for the Viersen
exercise, where the fire was set deliberately under controlled
conditions. Each folder exercises one
of four commander-level use cases (Table~\ref{tab:datasets}), presented in
turn below.

\begin{table}[t]
\caption{Field datasets and the SAR use case each exercises.}
\label{tab:datasets}
\centering
\footnotesize
\setlength{\tabcolsep}{4pt}
\begin{tabular}{@{}llll@{}}
\toprule
Use case & Incident & Pattern & Images \\
\midrule
ABC reconnaissance & Industrial hall, Berlin 2021 & meander & 167 \\
Flood & Erftstadt/Blessem 2021 & meander & 140 \\
Structural fire & Apartment building, Essen 2022 & meander & 159 \\
Vegetation fire & Exercise, Viersen 2025 & meander & 52 \\
\bottomrule
\end{tabular}
\end{table}

\begin{figure}[t]
\centering
\includegraphics[width=0.48\columnwidth]{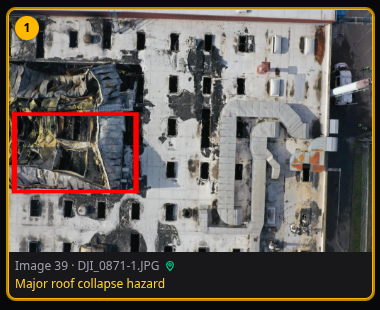}\hfill
\includegraphics[width=0.48\columnwidth]{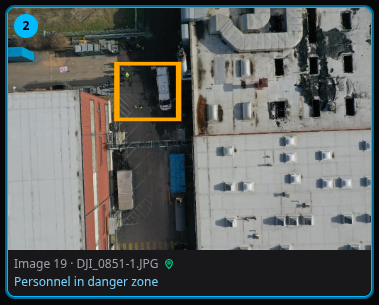}\\[2pt]
\includegraphics[width=0.48\columnwidth]{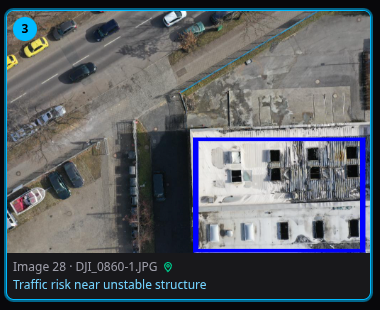}\hfill
\includegraphics[width=0.48\columnwidth]{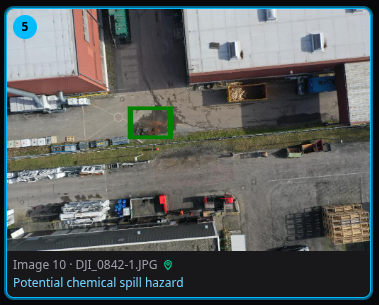}
\caption{Four of the shortlisted frames for the ABC-reconnaissance use case
(burned-out industrial hall, Berlin), with the model's rank, annotation, and
${\sim}$10-word justification: (1)~major roof collapse, (2)~personnel at
risk, (3)~traffic beside the unstable structure, (5)~the spill flagged as a
potential chemical hazard.}
\label{fig:industrial}
\end{figure}

The first use case is ABC (hazardous-substance) reconnaissance, exercised by
post-fire reconnaissance of a burned-out industrial hall in Berlin
(Fig.~\ref{fig:industrial}). The generic triage query already reasons from a
collapsed roof section down to a possible spill on the asphalt, and items
such as ``traffic risk near unstable structure'' appear alongside it; taken
together, the ranked hits read as a first draft of the situation report.
Notably, the spill hit emerged unprompted under the generic query.
Because task knowledge lives in the prompt, the same
folder can be re-tasked for dedicated ABC reconnaissance with a one-line
change, e.g.\
``Are there signs of leaking or hazardous substances (spills,
discoloration, drums, damaged tanks)? Mark them.'', where a fine-tuned fire
detector simply has no class to answer with. The honest boundary:
from the air a dark
patch may be oil, extinguishing water, or shadow; the shortlist directs a
ground team with measurement equipment to a location; it does not replace
them.

\begin{figure}[t]
\centering
\includegraphics[width=0.48\columnwidth]{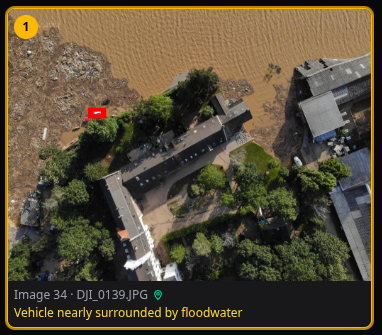}\hfill
\includegraphics[width=0.48\columnwidth]{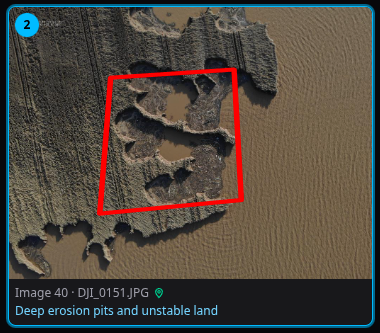}\\[2pt]
\includegraphics[width=0.48\columnwidth]{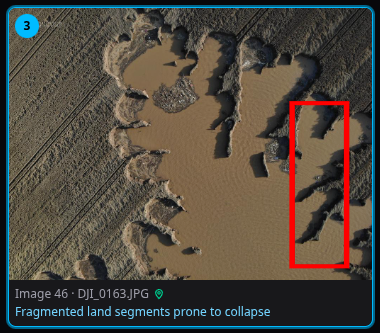}\hfill
\includegraphics[width=0.48\columnwidth]{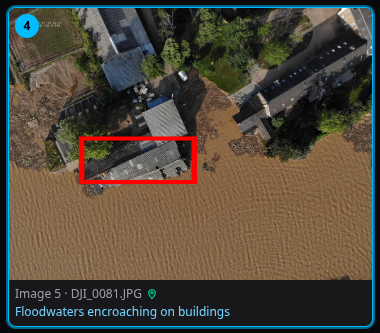}
\caption{Four of the shortlisted frames for the flood use case
(Erftstadt/Blessem), with the model's rank, annotation, and ${\sim}$10-word
justification: (1)~vehicle nearly surrounded by floodwater, (2)~deep erosion
pits and unstable land, (3)~fragmented land segments prone to collapse,
(4)~floodwater encroaching on buildings.}
\label{fig:floodframes}
\end{figure}

The second use case is the July 2021 Erftstadt/Blessem flood.
\blind{A deployment during this flood~\cite{surmann2022erftstadt}
supplies}{Our own deployment during this
flood~\cite{surmann2022erftstadt} supplies}
UAV imagery of destroyed buildings, infrastructure, and environment
around the Blessem gravel-pit erosion, where the operational
question was which structures were acutely threatened by the erosion edge.
The canonical triage query on the 140-image Erftstadt mission
(Fig.~\ref{fig:floodframes}) in batch mode answers it directly: the rank-1 hit flags a vehicle nearly surrounded by floodwater, and further hits
mark deep erosion pits and fragmented land prone to collapse at the erosion
edge, plus floodwater encroaching on
buildings.

\begin{figure}[t]
\centering
\includegraphics[width=0.48\columnwidth]{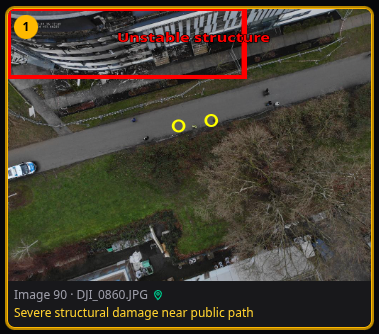}\hfill
\includegraphics[width=0.48\columnwidth]{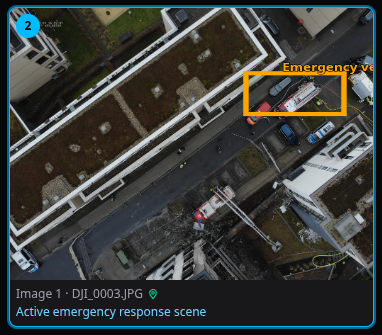}\\[2pt]
\includegraphics[width=0.48\columnwidth]{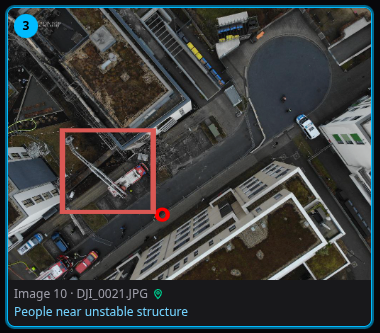}\hfill
\includegraphics[width=0.48\columnwidth]{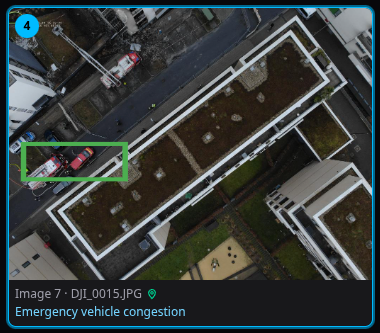}
\caption{Four of the shortlisted frames for the structural-fire use case
(Essen), with the model's rank, annotation, and ${\sim}$10-word
justification: (1)~severe structural damage near a public path, (2)~active
emergency response scene, (3)~people near the unstable structure,
(4)~emergency-vehicle congestion.}
\label{fig:essenframes}
\end{figure}

The third use case is a structural fire: post-fire reconnaissance of a
burned-out apartment building in Essen. Here
the value is not detection (the building is obviously damaged) but hazard
reasoning across the set: which parts of the structure are unstable,
who or what around it is exposed. Such items are not classes any pre-trained
detector emits; they are judgments about the relation between observations,
which is what a commander actually asks for. The batch-mode triage run on the 159-image Essen mission (Fig.~\ref{fig:essenframes}) surfaces
exactly such judgments:
the rank-1 hit marks severe structural damage adjacent to a public path, and
the summary flags civilians passing close to the debris of the collapsed
facade.

\begin{figure}[!htbp]
\centering
\includegraphics[width=0.48\columnwidth]{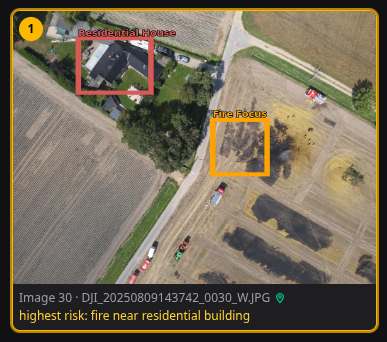}\hfill
\includegraphics[width=0.48\columnwidth]{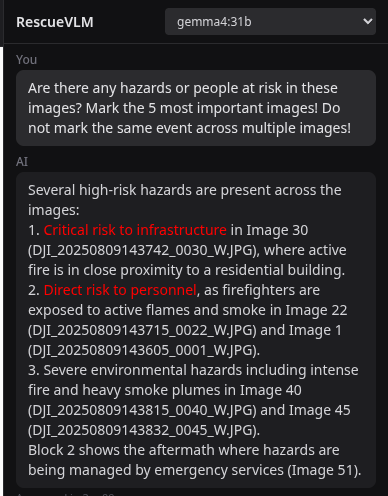}
\caption{Hazard prioritization by \code{gemma4:31b} in RescueVLM on the
Viersen dataset. Right: ranked summary of the most critical images,
distinguishing risk to infrastructure, risk to personnel, and environmental
hazards, with each finding referenced by image index and file name. Left:
the top-ranked result (Image~30), in which the model marks the active fire
and the adjacent residential building. The annotations are grounded in the
image content and directly linked to the textual assessment.}
\label{fig:gemma_shortlist}
\end{figure}

The fourth use case is a vegetation fire, captured in a 52-image meander
flight during a live fire exercise in Viersen: an active
surface fire over harvested fields, spreading near a road and adjacent farm
buildings, set deliberately but burning for real. The canonical triage query
returns a five-image shortlist (Fig.~\ref{fig:gemma_shortlist}): both distinct
active fire sources, ground personnel
inside the burned area, and the point where the burn approaches buildings. For the commander this is
the mission condensed to one screen: $52 \rightarrow 5$ frames
(${\approx}90\,\%$ fewer images viewed), each with a one-line justification and
a location. One observation cuts both ways: the shortlist includes the fire
crews already working the scene: known information, arguably noise; yet
suppressing it is a one-line prompt refinement (``ignore emergency personnel
already on scene''), not a retraining run, which is precisely the maintenance
model the tool argues for.

\subsection{Cross-cutting observations}
Two behaviors recur across the use cases. Exposure reduction: from a
52-image mission, the tool highlights a ranked, annotated shortlist of five
images, so the initial triage can rest on ${\approx}10\,\%$ of the frames
instead of a frame-by-frame screening of the entire set. The remaining images
are neither hidden nor discarded; they stay one click away for full review.
Deduplication: the contract's deduplication clause, which requires that each
distinct real-world event be reported only once even when it is visible in
several frames, held in the runs shown, with each fire source marked exactly
once despite appearing in dozens of overlapping frames.

\subsection{Toward quantitative evaluation}
Conventional metrics such as bounding-box overlap do not adequately capture RescueVLM's operational utility. 
SketchVLM~\cite{collins2026sketchvlm} evaluates performance on structured visual reasoning tasks, including maze 
navigation, ball-drop trajectory prediction, and object counting, and uses VLM judges to assess annotation quality 
and annotation--text alignment. These proxy measures cannot establish whether annotations support effective decisions 
in real rescue scenarios. We therefore argue that meaningful evaluation of RescueVLM's operational value requires metrics 
derived from field trials with search-and-rescue experts.

\section{Discussion}
\label{sec:discussion}
Every evaluation case in Sec.~\ref{sec:eval} is drawn from exactly the class of material that GDPR and BOS practice forbid sending to a foreign cloud API. This is also why
we report no cloud-frontier upper bound: the evaluation data itself cannot
lawfully be uploaded, which is not a gap in the evaluation but the core
argument for the on-premise design.

The human-factors payoff follows from the same numbers. Given the documented
psychological cost of imagery screening (Sec.~\ref{sec:intro}), viewing roughly
$10\,\%$ of frames is a concrete harm reduction, though residual exposure
remains: someone still sees the hard images, and the false-negative risk means
occasional full-set review cannot be abolished. Keeping the ranked shortlist
visibly provisional, with editable annotations and the full set one click
away, is our concrete implementation of appropriate
reliance~\cite{lee2004trust}.

For a SAR organization the relevant lifecycle question is not which model is
best today but what survives the next release. In RescueVLM the prompts,
output contract, parser, and interface are all model-independent, and
re-running the use-case queries of Sec.~\ref{sec:eval} becomes the acceptance
test that gates each new engine into service.

Several limitations remain (Sec.~\ref{sec:limitations}). The evaluation
presented is qualitative; a quantitative evaluation through field trials with
SAR experts is still pending, and until its numbers
exist the tool's claims should be read as demonstrated capability, not
measured accuracy. False negatives are the structural risk of any triage: an
event the model leaves off the shortlist is precisely the one nobody looks at,
which is why the tool ranks rather than filters and never deletes or hides an
image in batch mode. Off-the-shelf VLMs hallucinate objects~\cite{li2023pope} and forcing
JSON output can degrade reasoning~\cite{tam2024speakfreely}; both argue for
the conservative contract (ordering plus pointers, no diagnosis). Missions beyond
the 50-image batch limit are split into batches whose findings are merged in a
text-only synthesis step, so cross-batch deduplication and ranking rest on the
model's notes rather than on the images themselves. And triage quality
floats with the chosen engine, the price of having no weights of our own,
which re-running the per-dataset queries makes visible rather than
hides.

\section{Limitations}
\label{sec:limitations}


%


RescueVLM is model-agnostic at the architectural level, but its practical reliability remains dependent on the underlying VLM. SketchVLM reports its strongest performance with cloud-based frontier models, which are excluded by our privacy requirements, and demonstrates transfer to strong open-weight VLMs. However, several of these models, including \code{Qwen3.5-397B-A17B}, which performed well on SketchVLM~\cite{collins2026sketchvlm}, exceed the 32GB memory of our single RTX 5090 GPU.

\begin{figure}[!t]
\centering
\begin{minipage}{0.95\columnwidth}
\vspace*{6pt}
\centering
\includegraphics[height=3.2cm,width=0.48\linewidth,keepaspectratio]{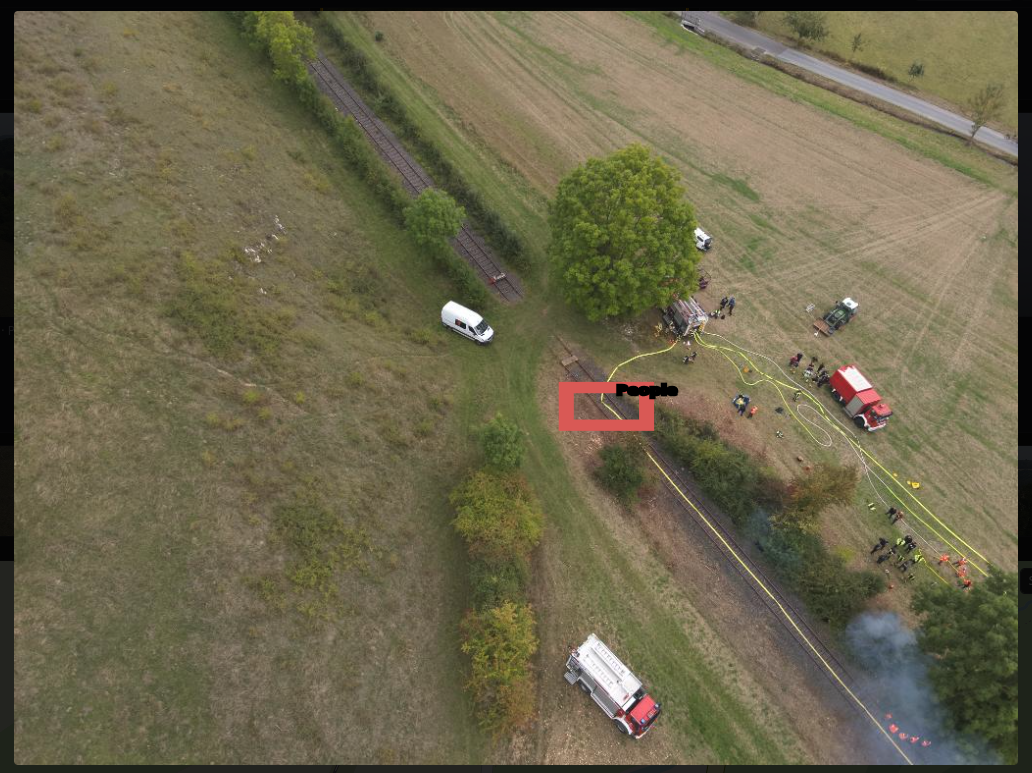}\hfill
\includegraphics[height=3.2cm,width=0.48\linewidth,keepaspectratio]{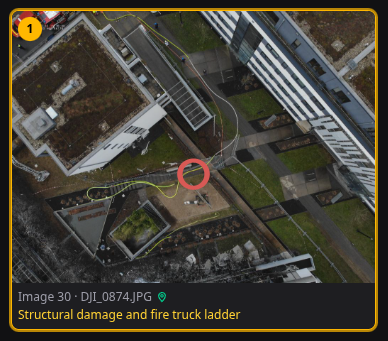}
\caption{Hallucinations by \code{nemotron3:33b}. Left: the model draws a
rectangle and labels it as people, although the marked region contains no
people and lies away from the center of the scenario. Right: the model
places a circle at an arbitrary location and describes it as structural damage
and a fire truck ladder, although there is no damage at that position and no
ladder in the image. In both cases, the annotations are confidently labeled
but not grounded in the image content.}
\label{fig:nemotron_hallucination}
\end{minipage}
\end{figure}

The locally deployable alternatives failed in qualitatively different ways.
\code{muse-glimmer:30b} aborted its analysis and defaulted to reporting that no
people at risk or life-safety hazards were discernible, effectively producing
false negatives. \code{nemotron3:33b} hallucinated events and annotated image regions
unrelated to the scenario (Fig.~\ref{fig:nemotron_hallucination}). \code{qwen3-vl:32b} was limited by its small
context window, which accommodated only about 25 images, and did not return
output in the required structure, so RescueVLM could not process its results.
\code{qwen3.8:27b} returned structured, indexed output, but its content was
noticeably less precise than that of \code{gemma4:31b}.

Among the locally deployable engines tested so far, only \code{gemma4:31b}
proved usable on the field data, producing indexed annotations with
accompanying notes. With identical prompts and the same output contract, none
of the other models achieved comparable reliability, as they struggled to
follow the framework's instructions and to use SketchVLM's annotation tools
correctly. The practical success of RescueVLM therefore depends on advances in
locally deployable open-weight VLMs, particularly their ability to follow
instructions and interact reliably with frameworks such as SketchVLM.

VLMs can hallucinate objects~\cite{li2023pope}, potentially diverting the attention of responders toward nonexistent hazards, while forcing structured output (JSON) can degrade reasoning~\cite{tam2024speakfreely}. These limitations are particularly critical in real-world rescue scenarios, where decisions can directly affect human safety. RescueVLM therefore presents rankings and annotations as provisional aids that require verification against the original imagery before operational action. Editable annotations and access to the full image set support human review but do not eliminate hallucinations or missed hazards.

\section{Conclusion}
We presented RescueVLM, an open-source, fully on-premise, training-free and
model-agnostic tool that triages folders of UAV survey imagery for
search-and-rescue. Keeping all task knowledge in prompts and a JSON+SVG output
contract answers three constraints at once: sensitive incident imagery stays on
the box (deployable under GDPR and BOS rules where cloud VLM services are not),
no fine-tuned weights depreciate on the fine-tune treadmill, and the screening
workflow is inverted so that fewer responders see fewer traumatic frames.
Across four use cases on German fire-brigade field data (vegetation fire,
structural fire, hazardous-substance reconnaissance, and flood), a single
natural-language query condenses a 52-image mission into a five-image
annotated, geo-anchored shortlist, and re-tasking the same imagery for a
different hazard class is a one-line prompt change rather than a retraining
run. The quantitative evaluation through expert field trials motivated in
Sec.~\ref{sec:eval} is future work; run per engine generation, it turns
``grows with the models'' from a slogan into an acceptance test.

\section{Future Work}
Future work
includes a first-class action-recommendation output
field and a field user study with
\blind{rescue-robotics}{DRZ} and fire-brigade personnel. Additionally, extending the system to support live video streams would be a valuable direction.

\ifblind\else
\section*{\raisebox{-2.5pt}{Acknowledgment}}
This work is funded by the Federal Ministry of Research, Technology and Space (BMFTR) under grant 13N16478 (E-DRZ), cf.\ \url{https://www.rettungsrobotik.de}.
\fi

\sloppy
\makeatletter
\g@addto@macro\@IEEEbibitemprefix{\rightskip 1.5pt\relax}
\makeatother
\bibliographystyle{IEEEtran}
\bibliography{bibliography}

\end{document}